\documentclass[runningheads]{llncs}
\usepackage[T1]{fontenc}
\usepackage{graphicx}
\usepackage[breaklinks]{hyperref}
\usepackage{tabulary}
\usepackage{wrapfig}
\usepackage{float}

\begin{document}
\title{Leveraging Large Language Models and Survival Analysis for Early Prediction of Chemotherapy Outcomes}
\titlerunning{Early Prediction of Chemotherapy Outcomes}
%
\author{
    Muhammad Faisal Shahid\inst{1}\and
    Asad Afzal\inst{1}\and
    Abdullah Faiz\inst{1}\and\\
    Muhammad Siddiqui\inst{1}\and 
    Arbaz Khan Shehzad\inst{1}\and
    Fatima Aftab\inst{1}\and\\
    Muhammad Usamah Shahid\inst{1}\and
    Muddassar Farooq\inst{1}
}
\authorrunning{F. Shahid et al.}
\institute{CureMD Research, 80 Pine St 21st Floor, New York, NY 10005, United States
\url{http://www.curemd.com}\\
\email{\{faisal.shahid, asad.afzal, abdullah.faiz, muhammad.siddiqui, arbaz.khan,\\
    fatima.aftab, muhammad.usamah, muddassar.farooq\}@curemd.com}}

\maketitle          
\begin{abstract}
\vspace{-0.3cm}
Chemotherapy for cancer treatment is costly and accompanied by severe side effects, highlighting the critical need for early prediction of treatment outcomes to improve patient management and informed decision-making. Predictive models for chemotherapy outcomes using real-world data face challenges, including the absence of explicit phenotypes and treatment outcome labels such as cancer progression and toxicity. This study addresses these challenges by employing Large Language Models (LLMs) and ontology-based techniques for phenotypes and outcome label extraction from patient notes. We focused on one of the most frequently occurring cancers, breast cancer, due to its high prevalence and significant variability in patient response to treatment, making it a critical area for improving predictive modeling. The dataset included features such as vitals, demographics, staging, biomarkers, and performance scales. Drug regimens and their combinations were extracted from the chemotherapy plans in the EMR data and shortlisted based on NCCN guidelines, verified with NIH standards, and analyzed through survival modeling. The proposed approach significantly reduced phenotypes sparsity and improved predictive accuracy. Random Survival Forest was used to predict time-to-failure, achieving a C-index of 73\%, and utilized as a classifier at a specific time point to predict treatment outcomes, with accuracy and F1 scores above 70\%. The outcome probabilities were validated for reliability by calibration curves. We extended our approach to four other cancer types. This research highlights the potential of early prediction of treatment outcomes using LLM-based clinical data extraction enabling personalized treatment plans with better patient outcomes.

\end{abstract}

\section{Introduction}

Chemotherapy is widely regarded as a core treatment for breast cancer, yet it
often poses significant physical, emotional, and financial burdens on patients.
The uncertainty of chemotherapy success compounds this stress, underscoring
the need for methods that can predict outcomes earlier in the treatment process.
Early prediction of potential treatment outcomes can not only guide clinical decisions but can also reduce unnecessary expenses and alleviate patient anxiety. In this
vein, recent research has explored the integration of biomarkers~\cite{ref2},
imaging-based insight~\cite{ref3}, and molecular characteristics of tumors~\cite{ref4}
to refine chemotherapy prediction and personalize therapy strategies.

While these studies mark substantial progress, they frequently highlight the
complexity of capturing the full clinical context. Some focus on a limited set
of biomarkers or imaging parameters, whereas others rely on machine learning
(ML) methods but omit critical patient details such as comorbidities, staging,
or performance scales~\cite{ref7}. Emerging approaches that employ Large
Language Models (LLMs) to structure information from clinical notes indicate
a promising direction, particularly to handle heterogeneous and unstructured
real-world data and entity extraction more systematically
\cite{ref10}.

Driven by these insights, our work aims to unify broader clinical features with
outcome modeling, ultimately providing clinicians with an early signal of whether
a given chemotherapy plan might fail. We draw on findings from earlier machine
learning studies on survival analysis~\cite{ref11}, focusing on robust data extraction
and meaningful feature engineering. By prioritizing both model interpretability
and reliable data integration, our goal is to create a framework that can help
improve breast cancer treatment and serve as a blueprint for similar predictive
tasks in other cancer types.

\section{Cancer Phenotype Extraction from Oncological Notes}
	\subsection{Introduction}
		AI-centered electronic health applications utilize big data collected from EHR systems. Coupled with AI/ML models, they are at the core of Real-World Evidence (RWE) paradigm. However, much of the valuable information required for building such applications is stored inside clinical notes, owing to the profession's legacy. For example, in our partner oncology EMR, 97\% of the oncologists record phenotypes in clinical notes. Furthermore, annotation of the phenotypes by expert oncologists would result in infeasible time and costs. Valuable RWE cannot be built around the true outcomes of effective cancer treatments.
		
		Building on the aforementioned issues, we propose a smart and autonomous framework for the extraction and annotation of cancer phenotypes utilizing a Retrieval-Augmented Generation (RAG) model and a Large Language Model (LLM) respectively. It runs using our on-premise, secure Nvidia A100 GPU clusters. We also perform a deep comparison with an earlier knowledge-driven system using the NCIt Ontology Annotator, while improving on the already existing ontology system.
		
	\subsection{Research Literature}
		Rule-based extractions are often reliable with a database housing relevant big data. The National Cancer Institute thesaurus (NCIt) \cite{golbeck2003national}, for instance, provides codes for over  171,000 classes and 500,000 relationships. Similarly, the Cancer Care Treatment Outcome Ontology (CCTOO) \cite{lin2018cancer} consists of a total of 1,133 classes.
		Towards deep learning, BioBERT \cite{lee2020biobert} and SciBERT \cite{beltagy2019scibert}, while boasting great F1 scores, extract limited entities. LLMs have been shown to be much more reliable for entity extraction. Huang et al. \cite{huang2024critical} used GPT-3.5 Turbo to extract lung and bone cancer information from pathological reports, showing the possibility of re-engineering an LLM for extracting information from provider notes with minimal human supervision.
		
	\subsection{Methodology}
\begin{figure}[!t]
		\centering
		\includegraphics[width=0.8\textwidth]{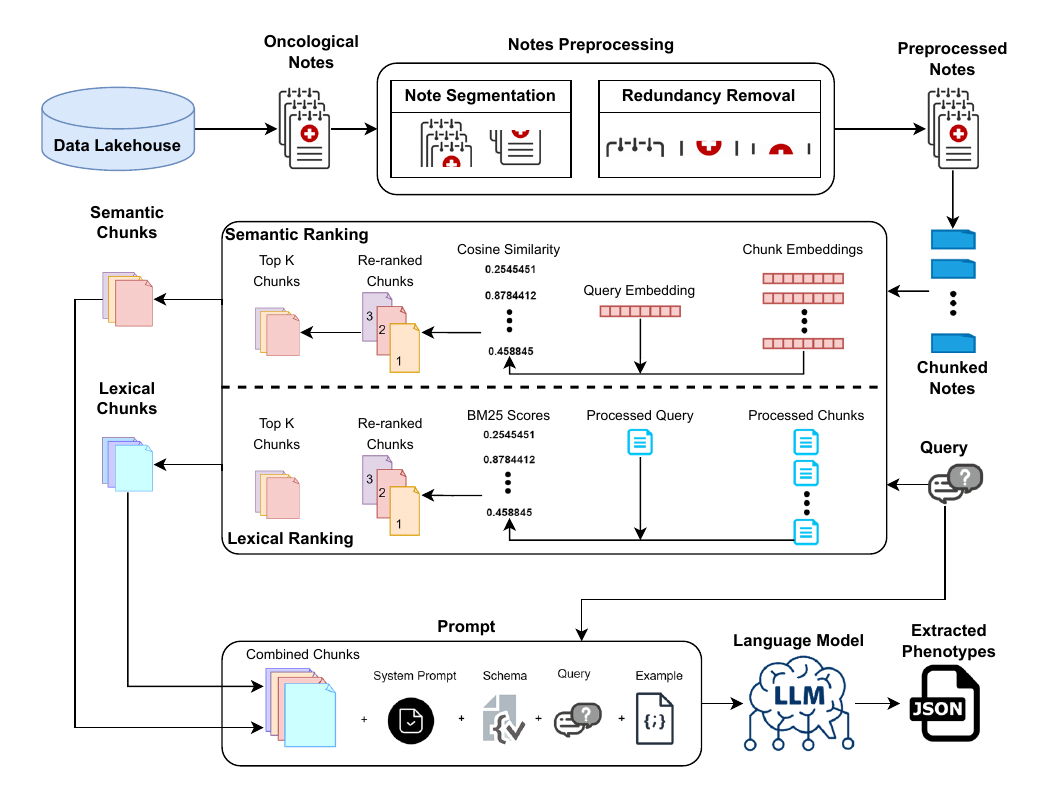}
		\caption{The LLM Annotation System's workflow. Both the lexical and semantic chunks are used to maximize information retrieval.}
		\label{fig:phenotype_workflow}
	\end{figure}

	The LLM Annotation System shown in Figure \ref{fig:phenotype_workflow} contains three main steps: (1) preprocessing notes to segment and remove redundancies; (2) computing cosine similarity and BM25 Scores to extract top $k$ note chunks both lexically and semantically; and (3) preparing a K-shot prompt to feed to the LLM for information extraction. The two well-known LLMs utilized for information extraction are LLaMA-3 8B and Mistral v0.2.
	
	The phenotypes important for breast cancer identification include TNM staging, stage group, tumor size, cancer grade, performance metrics, and biomarkers such as estrogen receptor (ER) and progesterone receptor (PR). To extract the information properly while minimizing the post-processing steps, we use Python's \texttt{Jsonschema} module to create a simple and effective schema that the LLM can adhere to.
	
	As LLMs are constrained by token limits, we apply a RAG system using the \texttt{mxbai} embeddings model that splits a note into chunks if the entire note exceeds 2,500 tokens. This ensures that the LLM has enough information to work with and maximize in-context retrieval ability. We find the top $k$ note chunks having the required information to be best at $k=10$.
	
	For the ontology system, primary improvements in the system included  updated regex for better biomarkers extraction, referencing for metastatic cancer, and processing improvements from the NCIt API.
	
	\subsection{Results and Discussion}
	A random sample of 150 admission and progress notes from different oncology practices was obtained to validate the results from both systems. Evaluating on five labels from each note, we evaluate on 750 labels in total, hand-annotated by a panel of five physicians headed by a senior resident oncologist in a partner university teaching hospital. Table \ref{tab:metrics} shows the metrics between the ontology system and the different LLMs used in our new system.
	
	The LLM Annotation system is rarely prone to (1) missing phenotypes due to vague semantics; and (2) hallucinating phenotypes not present in the clinical notes. The Ontology system's point of contention is the NCIt API which has been observed to fail annotation on some clinical notes, further driving the need for an improved phenotype annotation system.
	
	\begin{table}[h]
		\centering
		\begin{tabulary}{\textwidth}{|L|C|C|C|C|}
            \hline
			Model & Accuracy & Precision & Recall & F1-Score \\
			\hline
			LLaMA 3 8B & 86.13\% & 87.90\% & 94.95\% & 91.29\% \\
			\hline
			Mistral 7B v0.2 & 79.20\% & 81.35\% & 91.98\% & 86.34\% \\
			\hline
			Ontology & 85.04\% & 100.00\% & 83.80\% & 91.19\% \\
            \hline
		\end{tabulary}
		\caption{Classification metrics across both systems and the different LLMs. The Ontology system does not hallucinate and therefore achieves a perfect score in Precision.}
		\label{tab:metrics}
	\end{table}
	
	The system has since been updated to use better LLMs for phenotype extraction, including LLaMA 3.1 8B and Qwen 2.5 32B, combined with a critic agent that minimizes information that has been skipped or hallucinated by the LLM.
	
	\section{Cancer Labels Extraction from Oncological Notes}
	\subsection{Introduction}
	Chemotherapy treatment outcomes contain valuable information about the treatment itself and how it affects the cancer and the patient. They are vital to extract from clinical notes to effectively determine critical patient health and use them in decision-making processes. Multiple labels are extracted for treatment outcomes, categorized into three distinct branches: (1) progression; (2) toxicity; and (3) death/hospice. As with our LLM Annotation system, we once again use the power of LLMs to extract and interpret data to enhance the efficacy of cancer patient care pathways.
	\vspace{-10pt}
	\subsection{Research Literature} 
	Wang et al. \cite{wang2024entity} introduced an entity extraction pipeline for medical text records using LLMs. Despite the remarkable results, the authors note that the hallucination issues of LLMs require great attention. Monajatipoor et al. \cite{monajatipoor2024llms} also noted remarkable results in few-shot Named Entity Recognition (NER) for biomedical knowledge extraction using LLMs.
	
	The hallucination issues gave rise to thehe need of verifying results generated by the LLM, which is where our LLM Annotation system is refined not just for speedups and improvements in prompt engineering, but also an additional workflow after LLM generation that ensures that the generated response is valid and is contained in the note.
	\vspace{-10pt}
	\subsection{Methodology}
    We first set up a JSON schema to capture critical information from the notes. For each branch of the treatment outcomes, here is what's extracted as labels: (1) progression includes whether the cancer has progressed and details pertaining to the progression; (2) toxicity covers adverse effects, deterioration in quality life, and whether the treatment has been discontinued or modified; and (3) death/hospice covers if the patient has died or transferred to hospice, and date and other details on either.
    
    The Annotation system is updated for this objective to include a Critic agent at the end that validates the answer generated by the LLM to check whether the outcomes are correctly picked from the note and placed in the JSON structure. If any of the labels are generated incorrectly in the JSON structure, the output is returned to the LLM along with the note chunks to be reused for proper extraction. This feedback loop ensures that the LLM does not hallucinate.
	
	\begin{figure}
		\centering
		\includegraphics[width=0.8\textwidth]{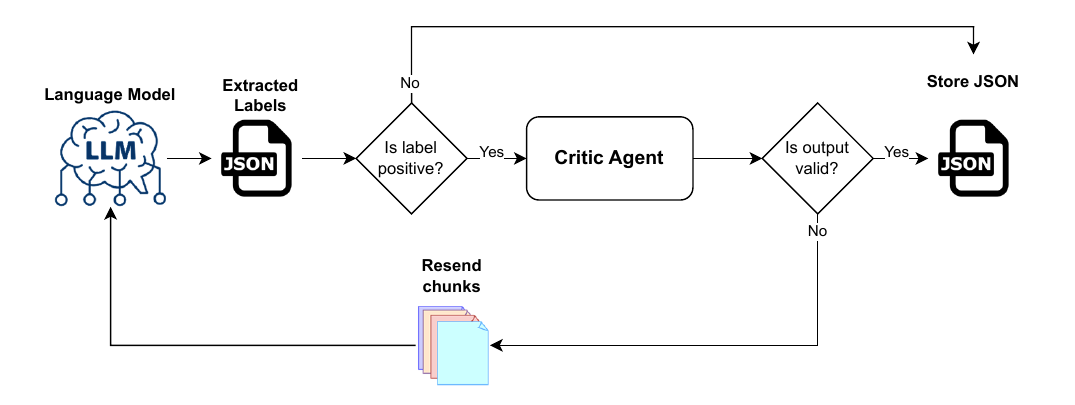}
		\caption{The step after LLM inference. The critic agent sends the note chunks back to the LLM if the generated JSON object is invalid.}
		\label{lab:critic_agent}
	\end{figure}
	
	The additional step in the system can be seen in Figure \ref{lab:critic_agent}. Our motivation with the feedback loop is to prevent any false alarms with the progression, toxicity, or hospice of death. As such, emphasis was given more so on correcting any false alarms.
	\vspace{-10pt}
	\subsection{Results and Discussion}
	A random sample of 225 oncological notes was obtained. Of those, all 225 notes were compared for progression and toxicity, and 50 from the 225 notes were compared for death/hospice. These notes were hand-annotated to check for the aforementioned labels if they were given in the note. The same notes are then passed to the system label by label and is validated with the original hand annotations.
	
	For this problem, only true and false positives were gathered to assess the reliability of the critic agent. Table \ref{tab:precision_table} shows the count of true positives, false positives, and the calculated precision across all three branches.
	
	\begin{table}[h]
		\centering
		\renewcommand{\arraystretch}{1.2}
		\begin{tabulary}{\textwidth}{|L|C|C|C|}
            \hline
			\textbf{Category} & \textbf{TP} & \textbf{FP} & \textbf{Precision} \\ \hline
			Progression & 179 & 46 & 79.5\% \\ \hline
			Toxicity    & 194 & 31 & 86.2\% \\ \hline
			Death       & 42  & 8  & 84.0\% \\ \hline
		\end{tabulary}
		\caption{Precision across Progression, Toxicity, and Death.}
		\label{tab:precision_table}
	\end{table}
	
	LLMs are generally prone to hallucinations and failing to understand information in the prompt. This necessitated the requirement of a critic agent with a feedback loop to provide stability and improvement in responses. The primary target was to reduce false positives as the LLM often had a tendency to declare a label to be present in the notes even if the information regarding it was absent. Most of the incorrect labels were what the LLM assumed given the details in the note. Much like how it was handled for the LLM Annotation system, our system focuses on explicit mention of progression with discontinuation, toxicity resulted in discontinuation, and death/hospice details to reduce such occurrences.
	
	Currently, improvements in the critic agent and the language model have vastly improved the results. The system runs in a parallel model to exponentially speed up the process on a single language model.
\vspace{-10pt}
\section{Breast Cancer Chemotherapy Outcome Modeling}
The breast cancer dataset consists of 3,409 patients, each with their first recorded treatment plan in the EMR. It includes important clinical features such as patient vitals, demographics, labs, comorbid conditions, and staging information, including overall stage, TNM staging, and cancer grade. Biomarker data includes ER, PR, and HER2 status, as well as performance metrics like ECOG and Karnofsky scores, all extracted using the Language Model.
\vspace{-10pt}
\subsection{Feature Vector}
The dataset comprises a variety of clinically relevant features to enhance predictive modeling. Key feature categories include:

\noindent\textbf{Vitals:} Body surface area (BSA).\\
\textbf{Demographics:} Age and gender.\\
\textbf{Labs:} Serum Creatinine (SrCr).\\
\textbf{Comorbid Conditions:} Diabetes, hypertension, kidney disease, and anemia, among others in ICD10 and elixhauser groups format along with readmission scores features.\\
 \textbf{Biomarkers:} Estrogen receptor (ER), Progesterone receptor (PR), and Human epidermal growth factor receptor 2 (HER2).\\
\textbf{Performance metrics:} Eastern Cooperative Oncology Group (ECOG) and Karnofsky performance status scores.\\
\textbf{Dosage and treatment Duration:} Standardized dosage per week and number of weeks in a treatment plan.\\
\textbf{Regimens and Combinations:} Chemotherapy treatments include both individual drugs and combinations. Examples include:
\begin{itemize}
    \item \textit{Individual drugs:} Docetaxel.
    \item \textit{Combination regimens:} Carboplatin+Docetaxel+Trastuzumab.
\end{itemize}

For regimen analysis, we compiled a list of unique chemotherapy drugs from the NCCN\cite{ref_url1} and NIH\cite{ref_url2} guidelines approved for breast cancer treatment, using drug GPI codes up to length 8. A total of 553 drugs and drug combinations were extracted, but most were supported by a single patient, indicating that these regimens were highly personalized. To ensure the robustness and generalizability of our feature vectors, we applied a patient support threshold of 20, reducing the number of regimen combinations used as features to 22. Additionally, we included 63 unique individual chemotherapy drugs as features, capturing a broad spectrum of treatment options while maintaining model reliability. 
The number of positive occurrences in the outcome labels in the cohort include progression with treatment discontinuation (42.12\%), treatment-related toxicity with discontinuation (25.01\%), and death or hospice care (2.11\%). By using these outcome labels, we establish the failure definitions used as targets in modeling and survival analysis.
The drugs with the highest failure percentages in our breast cancer cohort include Denosumab+Fulvestrant (79.6\%), Fulvestrant (70.7\%), and Carboplatin+Paclitaxel (72\%), indicating a substantial proportion of patients experiencing treatment failure with these regimens. On the other hand, Carboplatin+Docetaxel+Trastuzumab (30.5\%) and Zoledronic acid (34.5\%) exhibited the lowest failure percentages, suggesting better treatment responses among these patients.
\section{Survival Analysis and Modeling}
Survival modeling is a key statistical method for analyzing time-to-event outcomes in cancer treatment. In this study, we used survival analysis to predict chemotherapy treatment failures, defining time to failure from the start of chemotherapy to the observed event and incorporating censoring for patients without failure during the observation period. We employed Random Survival Forest (RSF) to calculate and visualize survival probabilities over time for both failure and non-failure groups by averaging the predicted survival functions. Model performance was evaluated using the concordance index (C-index), which measures how effectively the model ranks patients based on their risk scores.
\vspace{-15pt}
\subsubsection{Classification from Survival Models}
\begin{figure}[htbp]
    \centering
    \begin{minipage}[b]{0.48\textwidth} 
        \centering
        \includegraphics[width=\textwidth]{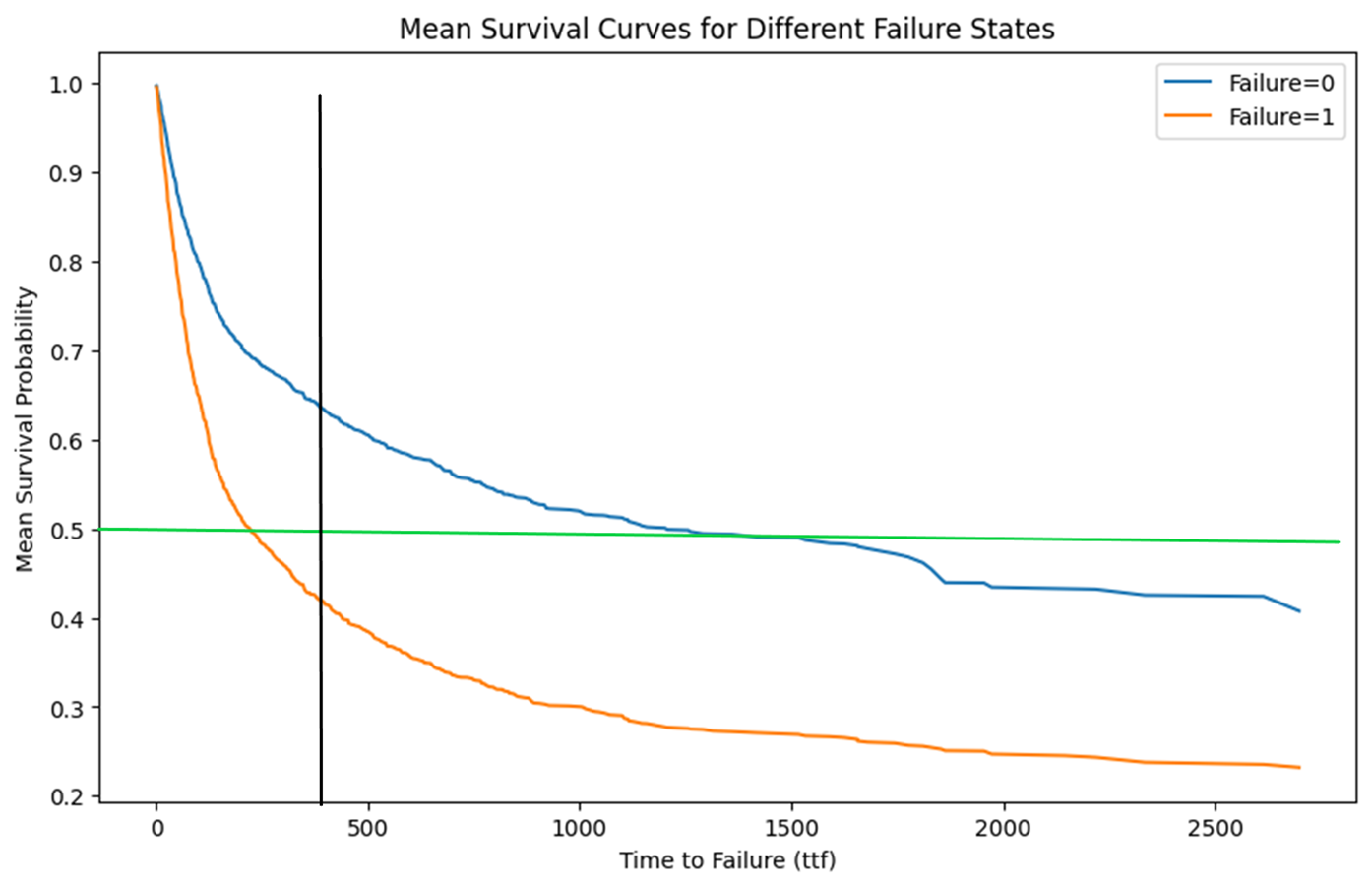}
        \caption{Mean survival curves for different failure states. The black vertical line marks the selected optimal time point for classification.}
        \label{fig:survival_curves}
    \end{minipage}
    \hspace{0.02\textwidth} 
    \begin{minipage}[b]{0.48\textwidth} 
        \centering
        \includegraphics[width=\textwidth]{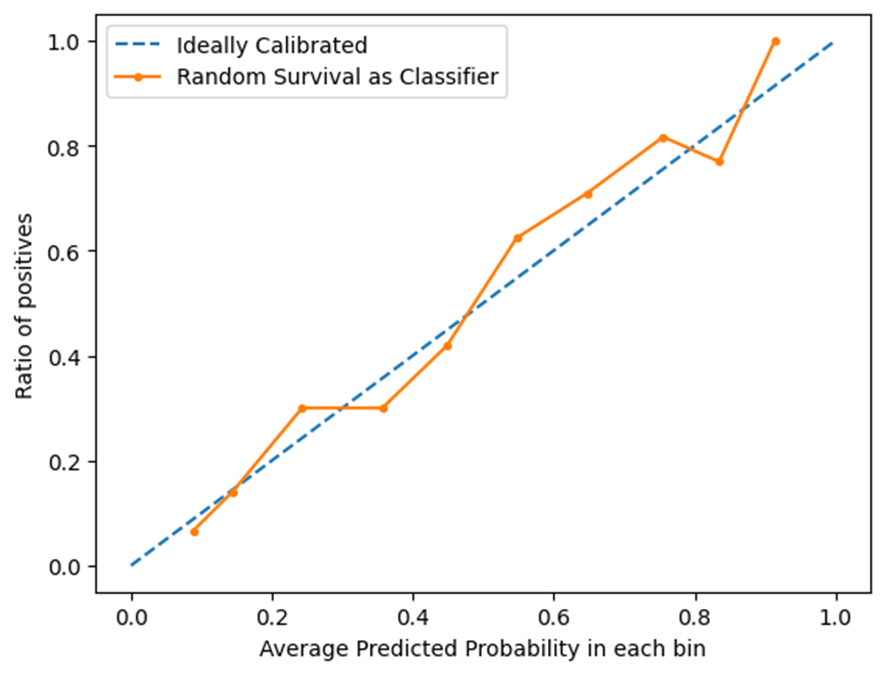}
        \caption{Calibration curve comparing the predicted probabilities with observed outcomes.}
        \label{fig:calibration_curve}
    \end{minipage}
\end{figure}

Beyond estimating survival probabilities, we employed RSF as a classifier to predict whether a patient would experience treatment failure. We strategically evaluated outcomes at different time points and selected the time point that yielded the optimal evaluation metrics, accuracy, and F1 scores for both classes separately. The survival curves presented in Figure \ref{fig:survival_curves} illustrate the mean survival probabilities for different failure states.

\vspace{-15pt}
\subsubsection{Calibration Curves}

To ensure confidence in our predictions, we evaluated our model's calibration by comparing the predicted probability distributions to the actual class distributions.We divided the predicted probability range into ten bins and computed the mean predicted probability within each bin. We then plotted this against the observed proportion of positive cases in that bin.

The resulting calibration curve, shown in Figure \ref{fig:calibration_curve}, demonstrates how well our model's predicted probabilities align with actual outcomes. The dashed blue line represents perfect calibration (i.e., an ideal classifier), while the solid orange line represents our model’s performance. A well-calibrated model should closely follow the diagonal, indicating that predicted probabilities accurately reflect observed frequencies.
\vspace{-10pt}
\section{Results}
For the Breast Cancer (C50) cohort, the RSF model achieved a C-index of 0.731, indicating a strong ability to differentiate between high- and low-risk patients. The classifier achieved an accuracy of 0.723 and an F1 score of 0.724 at the optimal time point of 431 days. The prevalence of treatment failure in this cohort was 50.3\%. These results highlight the effectiveness of the RSF model in predicting chemotherapy treatment failure. The most influential features in the model include weekly dose, overall stage, and metastatic stage (m stage), indicating the critical role in predicting survival outcomes. Among the treatment-related features, Denosumab and Cyclophosphamide+Doxorubicin HCL+Paclitaxel also showed significant importance, highlighting their potential impact on patient prognosis.
We further extended this approach to four additional prevalent cancer types, including Colon Cancer (C18), Lung Cancer (C34), Prostate Cancer (C61) and Multiple Myeloma (C90), by considering their relevant phenotypes extracted from the oncological notes with the same methodology. RSF demonstrated consistent performance across these cancer types, effectively capturing survival trends and treatment failure risks within each cohort.
Table \ref{tab:performance} summarizes the model's performance metrics across all cohorts, highlighting its predictive strength in classifying treatment failure.

\begin{table}[htbp]
\caption{Performance Metrics of RSF as a Classifier for Treatment Failure}
\centering
\begin{tabular}{|c|c|c|c|c|c|c|c|}
\hline
\textbf{Cohort} & \textbf{Cohort Size} & \textbf{C-index} & \textbf{Accuracy} & \textbf{F1 Score} & \textbf{Time Point} & \textbf{Failure} \\
\hline
C50 & 3409 & 0.731 & 0.723 & 0.724  & 431 & 0.503 \\
\hline
C18 & 1685 & 0.714 & 0.700 & 0.713 & 109 & 0.705 \\
\hline
C61 & 2079 & 0.757 & 0.731 & 0.678  & 238 & 0.496 \\
\hline
C90 & 1072 & 0.675 & 0.735 & 0.818  & 174 & 0.646 \\
\hline
C34 & 2366 & 0.660 & 0.677 & 0.766  & 122 & 0.609 \\
\hline
\end{tabular}

\label{tab:performance}
\end{table}

\section{Conclusion}
In this study, we leveraged Large Language Models (LLMs) and survival analysis for the early prediction of chemotherapy outcomes. By extracting phenotypes and treatment labels using a RAG and critic-agent loop, we improved the overall predictive accuracy. Random Survival Forest (RSF) achieved a C-index of 0.731 for breast cancer, with strong classification performance. The approach was also validated across four additional cancers, presenting our approach adaptability. By enabling early risk assessment, this framework supports personalized treatment planning. Future work will refine extraction methods,incorporate advanced survival models, undergo clinical validation, and extend applicability to additional cancer types.

\bibliographystyle{splncs04}
\bibliography{refs}

\end{document}